# DETECTION OF OBSTRUCTIONS IN OIL AND GAS PIPELINES: MACHINE LEARNING TECHNIQUES FOR HYDRATE CLASSIFICATION

Hellockston Gomes de Brito[1,*], Carla Wilza Souza de Paula Maitelli[2]; Osvaldo Chiavone-Filho[2]

[1]Doutorando em Engenharia Química, Universidade Federal do Rio Grande do Norte – UFRN;
[2]Docente do Programa em Engenharia Química, Universidade Federal do Rio Grande do Norte - UFRN

*e-mail do autor correspondente: hellockstongomes@gmail.com

**Abstract**

Oil and gas reserves are vital resources for the global economy, serving as key components in transportation, energy production, and industrial processes. However, oil and gas extraction and production operations may encounter several challenges, such as pipeline and production line blockages, caused by factors including sediment accumulation, wax deposition, mineral scaling, and corrosion. This study addresses these challenges by employing supervised machine learning techniques—specifically decision trees, the k-Nearest Neighbors (k-NN) algorithm (k-NN), and the Naive Bayes classifier method—to detect and mitigate flow assurance challenges, ensuring efficient fluid transport. The primary focus is on preventing gas hydrate formation in oil production systems. To achieve this, data preprocessing and cleaning were conducted to ensure the quality and consistency of the dataset, which was sourced from Petrobras' publicly available 3W project repository on GitHub. The scikit-learn Python library, a widely recognized open-source tool for supervised machine learning techniques, was utilized for classification tasks due to its robustness and versatility. The results demonstrate that the proposed methodology effectively classifies hydrate formation under operational conditions, with the decision tree algorithm exhibiting the highest predictive accuracy (99.99%). Consequently, this approach provides a reliable solution for optimizing production efficiency.

## Introduction

Oil and gas reserves are strategic assets that remain fundamental to the energy matrix, playing an essential role in transportation, energy generation, and industrial activities. As highlighted by Li et al. (2021), the continuous growth in demand, coupled with the limited availability of these resources, poses significant challenges to their management and exploitation. Nevertheless, a substantial increase in oil demand has been observed, reinforcing the urgency for developing more efficient and innovative extraction and management methods (Chen et al., 2024). In this context, deep and ultra-deepwater reserves emerge as a promising alternative to expand oil supply, mitigating concerns regarding its imminent scarcity (Nicolosi et al., 2024).

The transportation of multiphase fluids from offshore wells to processing facilities is a technically complex task, particularly due to challenges associated with maintenance, installation, and potential flow assurance issues (Obi et al., 2024). Despite technological advancements, oil and gas extraction operations remain subject to various challenges, including pipeline blockages—caused by paraffin deposition, scaling, etc., productivity losses, and equipment failures (Kumar, 2023).

This study focuses specifically on hydrate formation in oil production systems. Gas hydrates are crystalline solids composed of gas molecules resulting from the interaction between water and gas under high-pressure and low-temperature conditions. Also referred to as gas hydrates or clathrates, these structures form when water molecules arrange into cage-like structures that trap gas molecules, such as methane, ethane, or carbon dioxide, resulting in ice-like solids (Sadeh et al., 2024). The presence of these compounds in production pipelines can significantly reduce fluid flow and, in extreme cases, cause blockages (Liu et al., 2025). Thus, monitoring and preventing hydrate formation are critical aspects of oil and gas facility management and operations.

The pumping methodologies used in fluid transport operations are susceptible to operational failures and atypical conditions. Early detection of these anomalies is essential to avoid unplanned shutdowns, reduce corrective intervention costs, and consequently mitigate production losses (CID-Galiot et al., 2022). advancements in artificial intelligence, particularly supervised ML technologies have encouraged oil and gas companies to seek innovative solutions to optimize their processes.

In the field of Artificial Intelligence, Machine Learning enables pattern recognition through algorithms trained on expert-labeled datasets (Brasil, 2022). When applied to automated oil well systems, this technology facilitates the development of models capable of autonomously identifying abnormal operational conditions by analyzing historical and real-time data. Machine learning algorithms trained on time-series data from production wells can automatically classify operational states, identify anomalies, and support decision-making in automated oilfield environments (Li et al., 2022). In this context, the present study aims to apply Machine Learning techniques to data obtained from offshore production wells, focusing on the detection and classification of anomalies during oil and gas extraction, compared to normal operational conditions in these systems.

**Methodology**

The dataset utilized in this study originates from the publicly available 3W project developed by Petrobras, as detailed in Vargas et al. (2019). This dataset includes both real operational data from offshore wells and simulated instances manually constructed by domain experts to represent rare or underrepresented anomalous conditions. As highlighted by Krawczyk et al. (2017), such hybrid datasets are essential to overcome class imbalance, a common issue in supervised learning with real-world industrial data.

The dataset is composed of 1,025 labeled instances, categorized into three distinct operational classes: 597 normal operation cases, 344 instances reflecting rapid productivity loss, and 84 cases associated with hydrate formation. Each instance consists of time-series measurements from four process variables, the pressure and temperature variables are defined as follows:

- **P-TPT:** Pressure recorded by the transducer installed at the pump head (Pa).
- **T-TPT:** Temperature recorded by the transducer installed at the pump head (°C).
- **P-MON-CKP:** Upstream pressure at the production line's choke point (Pa).
- **T-JUS-CKP:** Downstream temperature at the production line's choke point (°C).

Methodological implementation was conducted using Python (version 3.10.12) and the scikit-learn library (version 1.4.2), selected for its comprehensive capabilities in classifier development. To ensure data quality and consistency, rigorous preprocessing was performed, including identification of missing values (NaNs), detection of frozen signals, and statistical outliers, followed by data normalization procedures.

Three established machine learning classifiers were employed: Decision Trees, the k-Nearest Neighbors (k-NN) algorithm (k-NN), and Naive Bayes classifier. These algorithms were chosen based on their demonstrated effectiveness in classification tasks, computational efficiency, and proven performance with high-dimensional datasets (Sheth, Tripathi, & Sharma, 2022). Figure 1 presents a flowchart of the methodology used for hydrate detection.

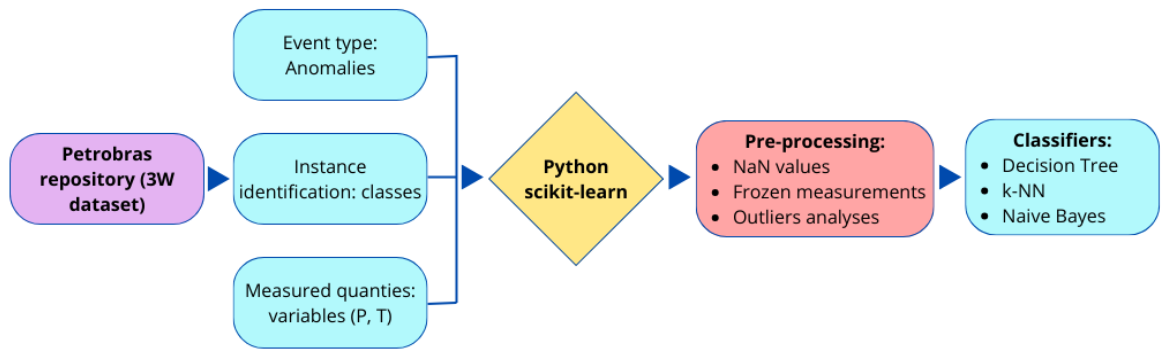

*Figure 1* - flowchart of the methodology.

For the statistical evaluation, we employed two non-parametric tests: the Kolmogorov-Smirnov (K-S) test and the Mann-Whitney U test. These methods are particularly valuable for comparing performance distributions among classification models (Decision Tree, k-NN, and Naive Bayes) when dealing with small sample sizes or unknown data distributions. The K-S test quantitatively assesses differences in distribution shapes, while the Mann-Whitney U test evaluates disparities in central tendency without requiring normality assumptions. The tests were performed at a 95% confidence level ($\alpha > 0.05$). Together, these tests provide robust, distribution-free comparisons of model effectiveness and performance consistency.

**Results and Discussion**

*Data Preprocessing and Quality Control*

When dealing with real-world datasets, issues related to data quality are both common and expected. Prior to initiating the preprocessing phase, a thorough assessment was conducted to identify and address key problems, including missing values (NaNs), frozen signals, and statistical outliers. The initial exploratory analysis revealed that, among the 8,200 variables evaluated across 1,025 instances, approximately 24.18% contained missing values (NaNs), while 9.94% of the entries corresponded to frozen or unchanging measurements, indicating potential sensor errors or logging issues. To address the issue of missing data, mean imputation was applied, wherein missing values in each column were replaced with the respective column mean. This method was selected for its simplicity and effectiveness in maintaining the statistical properties of the dataset. Outliers were identified through boxplot analysis (Figure 2), which enabled the identification of data points significantly deviating from the expected range. Identified outliers were subsequently treated to preserve the consistency and integrity of the dataset, thereby ensuring a more robust foundation for subsequent modeling and analysis (Mazarei et al., 2025).

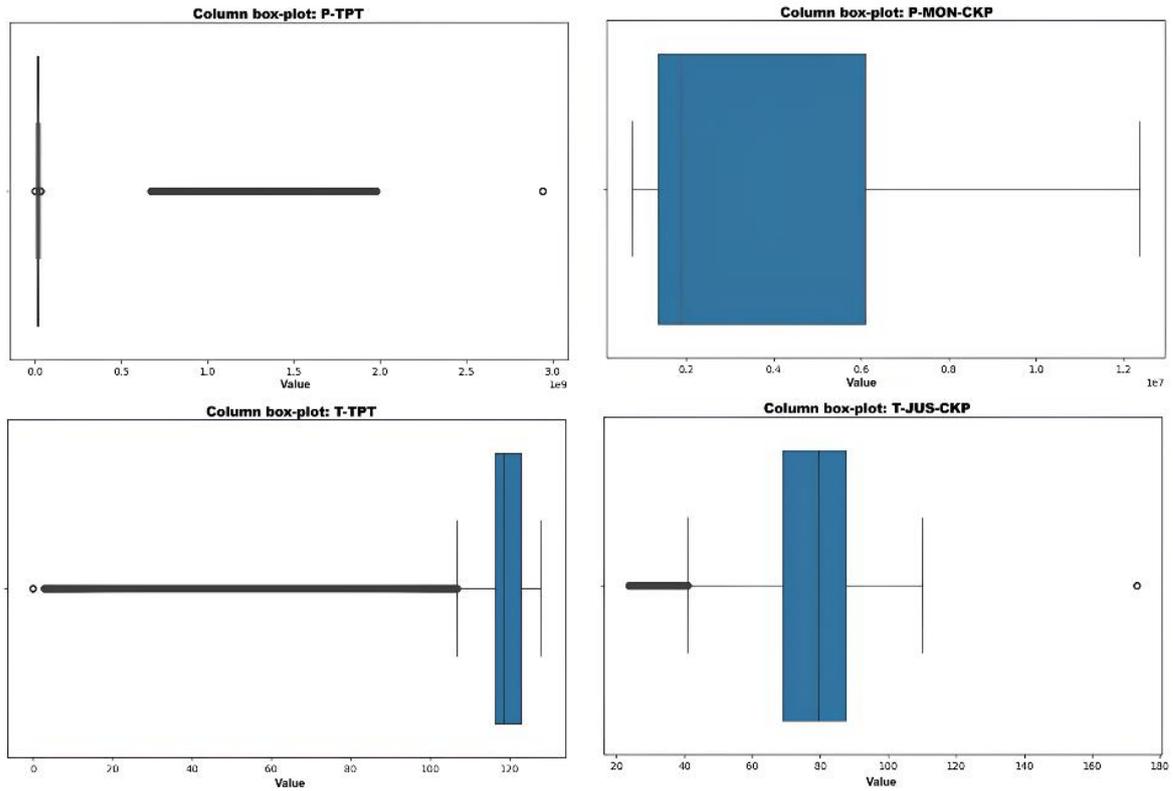

*Figure 2* - Boxplot of the variables in the dataset.

As illustrated by the boxplots, the variable **P-MON-CKP** was the only parameter that did not exhibit any outliers. Nonetheless, it demonstrated substantial variability and an asymmetric distribution, suggesting fluctuations in pressure that, while within the expected range, may reflect operational instabilities. In contrast, the **P-TPT** variable exhibited the highest proportion of outliers, accounting for 13.47% of its values. This indicates a considerable number of observations falling outside the expected range, coupled with a notable concentration of data points, which may suggest sensor anomalies or operational events.

The **T-TPT** variable presented 9.01% outliers, reflecting a moderate level of deviation from the expected thermal patterns. Similarly, **T-JUS-CKP** showed 6.36% of its values as outliers, indicating variability in downstream temperature measurements. These results highlight a combination of clustered and dispersed data behaviors across the dataset, reinforcing the importance of robust preprocessing to ensure accurate and reliable modeling.

*Classification Performance*

Table 1 presents the confusion matrix results obtained for each classification algorithm under the defined experimental conditions. These matrices offer a detailed evaluation of model performance, enabling analysis of classification accuracy, error distribution, and the ability of each method to distinguish between normal and anomalous operational states.

*Table 1* - Confusion matrix and Performace Metrics for each classifier agent. Where: Prod. Loss is Rapid production loss; Norm. cond. Is Normal well condition and F1-score is the Model evaluation metric.

| Classifiers | | Confusion Matrix | | | Performance Metrics | |
|---|---|---|---|---|---|---|
| | | Hydrates | P. prod. | Cond. Norm. | f1-score | Accuracy |
| **Decision Tree** | Hidratos | 67926 | 0 | 0 | 1.00 | 0.999 |
| | P. prod. | 0 | 298608 | 24 | 1.00 | |
| | Cond. Norm. | 0 | 25 | 397209 | 1.00 | |
| **k-NN** | Hidratos | 67896 | 0 | 30 | 1.00 | 0.997 |
| | P. prod. | 0 | 297682 | 950 | 1.00 | |
| | Cond. Norm. | 85 | 1088 | 396061 | 1.00 | |
| **Naive Bayes** | Hidratos | 1244 | 32462 | 34220 | 0.04 | 0.393 |
| | P. prod. | 0 | 298632 | 0 | 0.58 | |
| | Cond. Norm. | 329 | 396402 | 503 | 0.00 | |

Table 1 summarizes the classification performance of the **Decision Tree**, **the k-Nearest Neighbors (k-NN) algorithm (k-NN)**, and **Naive Bayes classifier** algorithms in identifying three operational states: **"Hydrates"**, **"Rapid Productivity Loss" (P. prod.)**, and **"Normal Condition" (Norm. cond.)**. The **Decision Tree** model exhibited exceptional performance, correctly classifying 67,926 instances as "Hydrates" with no misclassifications, 298,608 instances as "Rapid Productivity Loss" with only 24 errors, and 397,209 instances as "Normal Condition" with just 25 errors. With a perfect F1-score of 1.00 across all classes and an overall accuracy of 0.999, this model demonstrates an almost flawless ability to differentiate among operational conditions, making it highly reliable for real-time anomaly detection.

Similarly, the **k-NN** model also delivered high classification performance. It correctly identified 67,896 instances as "Hydrates" with 30 errors, 297,682 instances as "Rapid Productivity Loss" with 950 errors, and 396,061 instances as "Normal Condition" with 1,173 misclassifications. Despite a slightly higher error rate compared to the Decision Tree, the model maintained a strong F1-score of 1.00 across all classes and an overall accuracy of 0.997, confirming its robustness and suitability for practical deployment in production environments.

In stark contrast, the **Naive Bayes** classifier performed considerably worse, particularly in detecting instances of "Hydrates." It accurately classified only 1,244 instances in this category, resulting in 66,682 misclassifications. While it achieved perfect accuracy for the "Rapid Productivity Loss" class (298,632 correct classifications with no errors), its performance declined again in the "Normal Condition" category, with 396,402 correct classifications and 832 errors. The corresponding F1-scores were 0.04 for "Hydrates," 0.58 for "Rapid Productivity Loss," and 0.00 for "Normal Condition," leading to an overall accuracy of 0.393. These results indicate that the Naive Bayes model is unsuitable for this task in its current form.

The underperformance of the **Naive Bayes** model can be largely attributed to its assumption of conditional independence among features, which is unrealistic in this context. The dataset contains high-dimensional, correlated variables, particularly pressure and temperature measurements captured over time, that inherently violate the independence assumption (Yi et al., 2024). As a result, the model fails to capture complex interdependencies and misclassifies a substantial number of instances, especially within the minority class (Hydrates), where accurate identification is most critical.

In summary, both the **the Decision Tree algorithm** and **k-NN** classifiers exhibit excellent predictive performance and are well-suited for operational anomaly detection in oil and gas production wells. In contrast, **Naive Bayes** demonstrates significant limitations under the current dataset conditions and should be reconsidered or re-engineered if used in future analyses. Figures 3, 4, and 5 illustrate the classification outcomes for each method, emphasizing the contrast in predictive accuracy and class separability across the models.

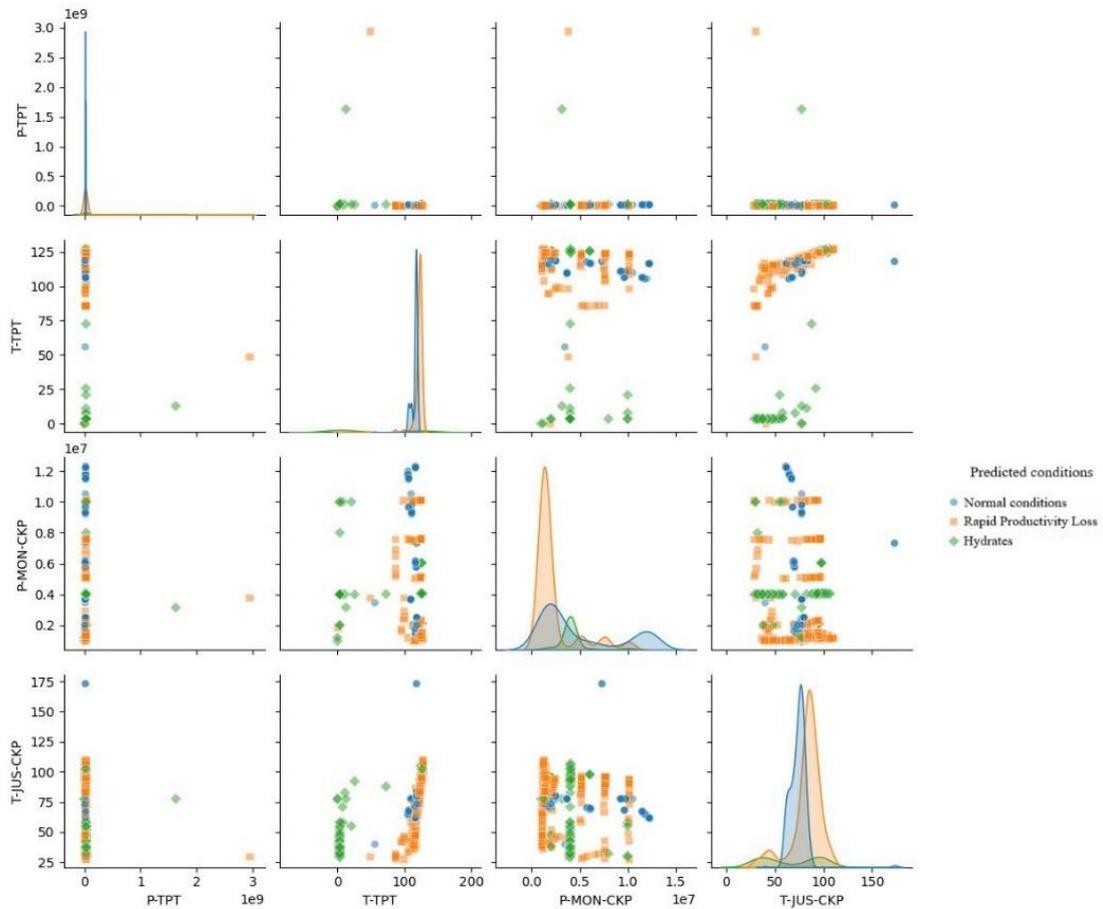

*Figure 3* - the Decision Tree algorithm algorithm for anomaly classification. Color-coded nodes represent: Blue circles - Normal conditions; Orange squares - Rapid productivity loss; Green diamonds - Hydrate formation.

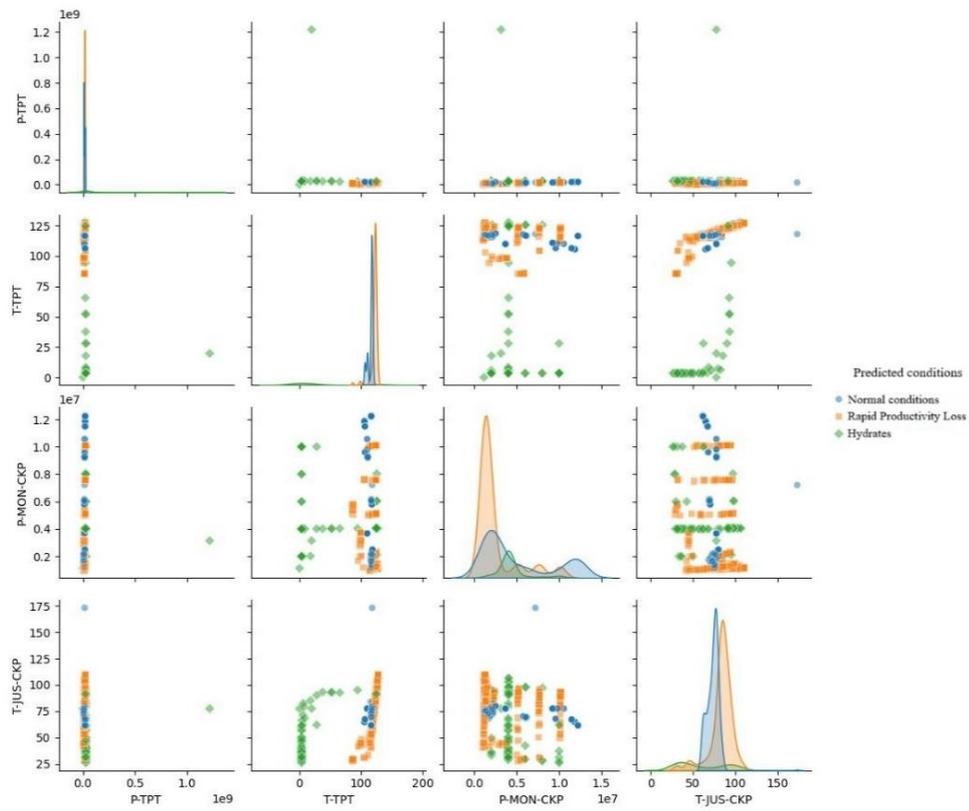

*Figure 4* - k-NN classification results for operational anomalies. Color-coded nodes represent: Blue circles - Normal conditions; Orange squares - Rapid productivity loss; Green diamonds - Hydrate formation.

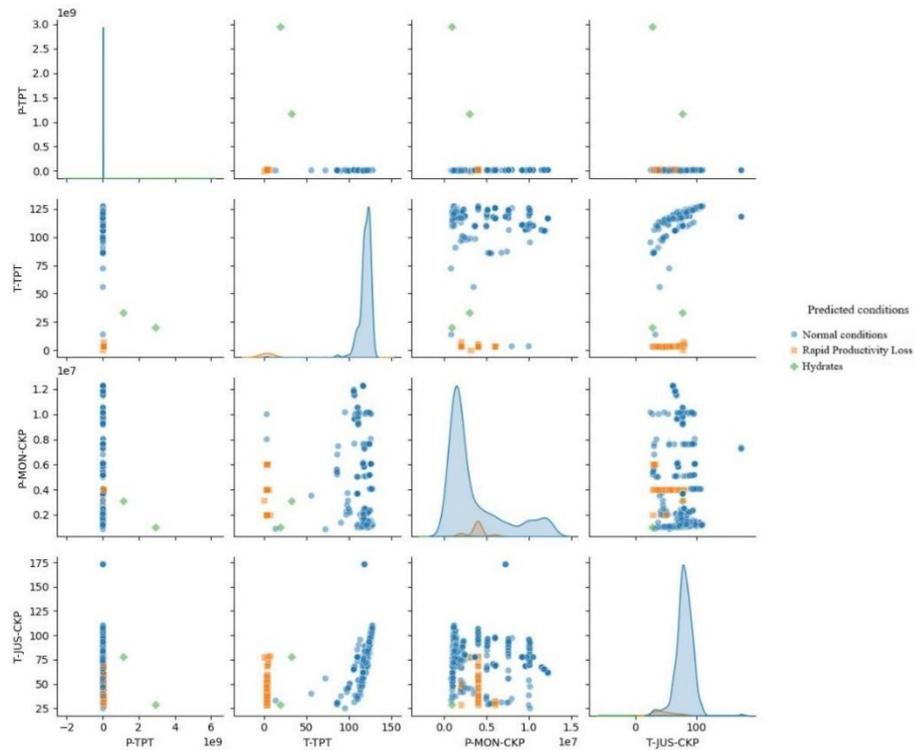

*Figure 5* - Naive Bayes classification results for operational anomalies. Color-coded nodes represent: Blue circles - Normal conditions; Orange squares - Rapid productivity loss; Green diamonds - Hydrate formation.

As illustrated in Figures 3, 4, and 5, a clear and consistent relationship emerges between the measured operational variables and the occurrence of failure events, particularly hydrate formation. Figure 5, in alignment with the performance metrics presented in Table 1, reveals considerable challenges related to high data dimensionality, especially given the underrepresentation of hydrate instances compared to other operational states. This class imbalance exacerbates the classification difficulty and emphasizes the need for models capable of handling sparse and non-linear feature distributions.

The formation of hydrates is typically associated with low-temperature and high-pressure conditions (Sadeh et al., 2024), a phenomenon clearly reflected in the cross-plots of **T-TPT** (temperature at the pump head) versus **P-MON-CKP** (pressure at the choke point) presented in Figures 3 and 4. These figures reveal that Hydrate cases were consistently associated with these extreme thermodynamic conditions, thus reinforcing the physical validity of the classification labels and highlighting the complex nature of hydrate detection.

Moreover, the confusion matrices for both the **Decision Tree** and **k-NN** classifiers exhibit highly similar classification patterns (as shown in Table 1), which is visually corroborated by the analogous clustering and distribution trends observed in their corresponding graphical outputs. This visual alignment serves to validate the high predictive consistency between the two models and their effectiveness in distinguishing between normal and anomalous operational conditions (Zhang et al., 2024).

A more detailed inspection of the **T-TPT** and **T-JUS-CKP** (downstream temperature) distributions under hydrate-forming scenarios further substantiates the thermal characteristics of hydrate presence. Under these conditions, data points for both temperature variables are predominantly localized within the 0–50°C range, demonstrating strong clustering in regions that favor hydrate crystallization. These findings not only support the thermal thresholds known to induce hydrate formation but also indicate that temperature variables serve as robust indicators for early detection (Sadeh et al., 2024).

Lastly, the diagonal dispersion plots, as initially introduced in Figure 2, effectively illustrate the variability and concentration of individual parameters, particularly highlighting the high-density clustering observed for the **P-TPT** (pressure at the pump head) variable. This reinforces the need for models that can effectively manage both concentrated and dispersed data distributions in order to achieve reliable classification.

In summary, the graphical analyses complement the quantitative results by offering a visual confirmation of the data patterns underlying classification performance. Together, they underscore the importance of temperature and pressure variables as key predictors in identifying hydrate formation and demonstrate the coherence between model output and the physical realities of offshore oil production environments.

*Statistical Analysis of Classifier Performance*

In order to rigorously evaluate the differences in performance between the classification algorithms employed in this study, namely **Decision Tree**, **k-NN**, and **Naive Bayes (NB)**, two robust, non-parametric statistical methods were applied: the **Kolmogorov-Smirnov (K-S) Test** and the **Mann-Whitney U Test**. These techniques were chosen due to their suitability for small sample sizes and their ability to assess distributional differences without the assumption of normality, which is especially relevant given the limited number of F1-score values available (i.e., one per class per model).

*Kolmogorov-Smirnov Test*

The **Kolmogorov-Smirnov Test** is a non-parametric method that compares the empirical distribution functions of two independent samples to determine whether they originate from the same continuous distribution (Table 2). The test is sensitive to differences in both location and shape of the empirical cumulative distribution functions (ECDFs) of the two samples.

*Table 2* – Results for Kolmogorov-Smirnov Test

| Comparison | KS Statistic | p-value |
|---|---|---|
| Decision Tree vs. k-NN | 0.00 | 1.000 |
| Decision Tree vs. NB | 1.00 | 0.100 |
| k-NN vs. NB | 1.00 | 0.100 |

The results of the K-S test indicate that there is no statistically significant difference between any of the classifier distributions at the 95% confidence level ($\alpha > 0.05$). However, important nuances can be observed:

The comparison between the Decision Tree algorithm and k-NN yielded a KS statistic of 0.00 and a p-value of 1.000, clearly indicating identical distributions of F1-scores across the three operational classes. This is consistent with the earlier descriptive observation that both models yielded near-perfect or perfect F1-scores in all categories.

Conversely, the comparisons involving Naive Bayes returned a KS statistic of 1.00, which is the maximum possible value, suggesting that there are substantial differences in the empirical distributions. However, with a p-value of 0.100, this difference does not reach statistical significance under conventional thresholds. Nonetheless, the high KS statistic value warrants attention, particularly given the practical implications of underperformance in the hydrate classification class.

*Mann-Whitney U Test*

The **Mann-Whitney U Test** is another non-parametric test used to assess whether two independent samples originate from the same distribution, specifically focusing on differences in medians. Unlike the K-S test, which compares entire distributions, the Mann-Whitney test evaluates whether one sample tends to yield higher values than another (Table 3).

*Table 3* - Results Mann-Whitney U Test

| Comparison | U Statistic | p-value |
| --- | --- | --- |
| Decision Tree vs. k-NN | 4.5 | 1.000 |
| Decision Tree vs. NB | 9.0 | 0.064 |
| k-NN vs. NB | 9.0 | 0.064 |

The comparison between Decision Tree and k-NN produced a p-value of 1.000, confirming that there is no significant difference in the central tendency (i.e., medians) of the F1-score distributions. This validates the assumption that both models perform equivalently in terms of their classification capacity across the operational classes.

The comparisons between Naive Bayes and both Decision Tree and k-NN produced p-values of 0.064, which, although not below the conventional threshold of 0.05, approach the significance boundary. This indicates a trend toward statistical significance, suggesting that, with a larger dataset or more classification categories, the difference in median F1-scores between Naive Bayes and the other models could indeed be confirmed as statistically meaningful.

Importantly, these results highlight that while Naive Bayes demonstrated visibly lower F1-scores—particularly for the hydrate class, where performance was critically inadequate—the small sample size (only three F1-score values per model) limits the statistical power of these tests. Nonetheless, the consistent underperformance of Naive Bayes across multiple metrics (e.g., accuracy, F1-score, confusion matrix misclassifications) supports the conclusion that it is unsuitable for the classification of operational conditions in this context.

The Kolmogorov-Smirnov and Mann-Whitney U tests statistically validate our empirical findings, confirming that Decision Tree and k-NN models exhibit indistinguishable classification performance ($\alpha > 0.05$), both demonstrating high reliability across operational conditions. While Naive Bayes shows statistically distinct performance patterns, particularly in hydrate detection, these differences lack significance at the 95% confidence level. Given the limited sample size, these results should be interpreted cautiously, warranting further validation with larger, more diverse datasets to strengthen statistical power and confirm these preliminary observations.

**Conclusion**

This study demonstrated the effectiveness of supervised machine learning techniques for detecting hydrate formation and other operational anomalies in offshore oil and gas production systems. Among the evaluated models, the **Decision Tree** classifier exhibited the best overall performance, achieving an F1-score of 1.00 and an accuracy of 99.9%, making it highly suitable for real-time anomaly detection under production conditions. The **k-NN** model also delivered robust results, while the **Naive Bayes** classifier showed significant limitations, particularly in scenarios involving complex, correlated features.

The consistent results underscore its robustness and reliability in the analyzed context. The findings confirm the hypothesis that machine learning techniques are effective in predicting hydrate-related failures in offshore environments. The models' performance validates the study's core premise, showing that even classical algorithms can deliver excellent predictive capability when properly tuned and applied to critical operational scenarios.

From a practical standpoint, the results highlight the potential for implementing the Decision Tree model in real-time monitoring systems, with direct implications for reducing unplanned downtime and optimizing production. Theoretically, this study advances the integration of AI techniques with well engineering, demonstrating the value of data-driven modeling in complex industrial operations. However, the research has limitations. Class imbalance and the restricted dataset size may affect the generalizability of the results. Nevertheless, the use of non-parametric statistical tests (Kolmogorov-Smirnov and Mann-Whitney) ensured robust analysis despite these constraints.

Future research should validate the models using larger datasets and explore more advanced architectures, such as deep neural networks, along with resampling techniques to address class imbalance. In summary, this work establishes a solid foundation for applying supervised classification algorithms in offshore monitoring applications, enhancing operational efficiency, safety, and automated decision-making.

**Acknowledgments**

The authors gratefully acknowledge Petrobras, the 3W Project, and Vargas et al. (2019) for providing access to the 3W Dataset used in this study. This research was financially supported by the Human Resources Program of the Brazilian National Agency of Petroleum, Natural Gas and Biofuels (PRH-ANP), funded through investments from qualified oil companies under Clause P,D&I of ANP Resolution No. 50/2015. The authors also acknowledge the Federal University of Rio Grande do Norte (UFRN) and the Automation Laboratory (LAUT) for their institutional support.